\documentclass[10pt,twocolumn,letterpaper]{article}

\usepackage[pagenumbers]{cvpr} % To force page numbers, e.g. for an arXiv version

   \definecolor{DBpurple}{RGB}{0,0,0}
   \newcommand{\DB}[1]{}
   
\definecolor{cvprblue}{rgb}{0.21,0.49,0.74}
\usepackage[pagebackref,breaklinks,colorlinks,allcolors=cvprblue]{hyperref}
\usepackage{multirow}

\def\confName{CVPR}
\def\confYear{2027}

\title{Reconstructing 3D+t Cardiovascular Flow From Sparse Planar Velocity Fields With Neural Fields}

\author{Viraj Patel\\
Department of Computer Science\\
University of Bath\\
{\tt\small vbp24@bath.ac.uk}
\and
Deblina Bhattacharjee\\
Department of Computer Science\\
University of Bath\\
{\tt\small db2466@bath.ac.uk}
\and
Katharine Fraser\\
Department of Mechanical Engineering\\
University of Bath\\
{\tt\small khf27@bath.ac.uk}
}

\DeclareMathOperator*{\argmin}{arg\,min}

\begin{document}
\maketitle
\begin{abstract}
Reconstructing time-resolved three-dimensional (3D+t) cardiovascular flow from sparse planar measurements is challenging because each plane observes only the two in-plane velocity components, on a small part of the volume. We propose a plane-aware neural-field framework that learns a continuous 3D velocity field from space and time coordinates through a per-plane projection, and then investigate the potential benefits of physics constraints. We evaluate the framework on computational aortic and aneurysm datasets, and on planes resampled from a real cardiac 4D Flow MRI scan. A plain multi-layer perceptron (MLP) was more accurate than classical radial basis function (RBF) interpolation on both computational datasets. For the aorta, a physics-constrained SIREN (PC-SIREN) reduced the rRMSE further to $0.3756$, compared with $0.4273$ for the plain MLP and $0.7256$ for RBF, and it was the most accurate model in every anatomical segment of the MRI data. For the aneurysms the plain MLP remained the most accurate model, with an rRMSE of $0.07$ to $0.16$ against $0.55$ to $0.71$ for PC-SIREN. Further experiments investigated scanning configuration, number of scans, and spatial resolution. These findings show that continuous neural fields can reconstruct volumetric cardiovascular flow from sparse planar measurements, and that physics constraints have to be matched to the flow.
\end{abstract}    
\section{Introduction}
\label{sec:intro}

Cardiovascular disease remains the leading cause of death globally, highlighting the need for accurate and accessible medical imaging tools \cite{chongGlobalBurdenCardiovascular2025}. Although magnetic resonance imaging (MRI) and computed tomography (CT) are commonly used for diagnosis due to their 3D imaging capabilities, echocardiography is a faster and less invasive alternative, while also being accessible to patients with medical implants. Significant advances in ultrasound hardware and computational methods have enabled capable 3D and time-resolved imaging \cite{fensterThreeDimensionalUltrasoundImaging2000, huangReviewRealTime3D2017}. Despite these advances, 3D ultrasound remains susceptible to artifacts such as stitching, blooming, and signal dropout, particularly when volumetric information is reconstructed from sparse observations \cite{saricRecommendationsIdentificationMitigation2026}. The larger size of matrix-array transducers can also increase susceptibility to acoustic shadowing, while the greater number of elements introduces a more pronounced trade-off between spatial and temporal resolution \cite{langEAEASERecommendations2012}.

As an alternative to relying solely on raw anatomical images, blood flow dynamics provide a sensitive indicator of cardiovascular abnormalities, motivating diagnostic approaches based on haemodynamic measurements for conditions such as aortic coarctation and heart failure \cite{secchiMRImagingAortic2009, zhongAssessmentCardiacDysfunction2016}. Blood velocity is clinically assessed using Doppler ultrasound, although this only measures a single velocity component along the ultrasound beam. Many techniques have been developed to recover 2D velocity fields, including vector flow mapping (VFM), blood speckle tracking (BST), and echocardiographic particle image velocimetry (EchoPIV) \cite{forsbergClinicalApplicationsUltrasound1998,nyrnesBloodSpeckleTrackingBased2020, avesaniVectorFlowMapping2021}. Current 3D ultrasound flow imaging, such as 3D colour Doppler, is typically limited to measuring a single component of the velocity field, while approaches such as 3D EchoPIV aim to recover three-dimensional flow information but remain constrained by the trade-off between spatial and temporal resolution \cite{chaouiThreedimensional3D4D2004, voorneveld4DEchoParticleImage2020}. Consequently, 2D techniques can provide more accurate and temporally resolved flow measurements, but these planar observations do not fully capture the inherently three-dimensional dynamics of cardiovascular flow.

These limitations raise the question of whether the accuracy of planar velocimetry can be combined with the volumetric representation of 3D flow by reconstructing the unobserved velocity field from sparse planar measurements. We propose a neural field framework that represents cardiovascular flow as a continuous function of space and time, enabling reconstruction of transient 3D velocity fields from a limited number of 2D observations. In this work, the planar velocity fields are sampled from computational simulations and from a 4D Flow MRI scan, standing in for the output of 2D ultrasound velocimetry, so that each reconstruction can be evaluated against a known 3D reference. The resulting representation provides continuous access to the reconstructed flow field throughout the volume and over time, while providing a potential alternative to dense volumetric flow acquisition that is used in 3D ultrasound. We systematically evaluate the influence of measurement density and scanning configuration on reconstruction accuracy and computational cost, and further investigate robustness to increasingly complex flow geometries. Our framework is benchmarked against a classical interpolation baseline. Furthermore, the continuous representation also enables the incorporation of physics-based constraints.

In summary, our contributions are
\begin{itemize}
    \item \textbf{Physics-free 3D+t reconstruction from planes:} We propose a plane-aware neural-field framework that reconstructs time-resolved 3D cardiovascular velocity fields from two-component measurements on freely oriented planes. Unlike prior learning-based reconstructions from planes, it requires no governing equations, boundary conditions, or pressure data, and the network is interchangeable.
    \item \textbf{Acquisition design:} We compare four handheld-probe scanning patterns and scan counts from 4 to 100, and identify two-axis rotation with 20 scans as an accurate and practical protocol.
    \item \textbf{Validation across flows and on real data:} Across aortic coarctation, intracranial aneurysms, and real whole-heart 4D Flow MRI, the framework reduces the rRMSE of RBF interpolation by 13 to 89\%, and we show where physics constraints add further accuracy.
\end{itemize}
\section{Related Work}

\paragraph{Implicit Neural Representations}

Neural fields encode a continuous signal as a network mapping spatial coordinates, and optionally time, to a quantity of interest such as colour, density or velocity \cite{xieNeuralFieldsVisual2022, garziaNeuralFieldsContinuous2025, patelDynamicReconstructionUltrasoundDerived2025}. NeRF~\cite{mildenhallNeRFRepresentingScenes2021} learns a continuous scene representation from posed 2D images by optimising a photometric loss through differentiable volume rendering, and has since been adapted to cross-sectional imaging \cite{sawdayeeOReXObjectReconstruction2023}. In medical imaging, MedNeRF~\cite{corona-figueroaMedNeRFMedicalNeural2022} learns continuous CT representations from sparse projections with an adversarial objective. For ultrasound, Ultra-NeRF~\cite{wysockiUltraNeRFNeuralRadiance2024} models planar measurements through tissue-dependent attenuation, reflectance, and scattering, and UlRe-NeRF~\cite{guoUlReNeRF3DUltrasound2024} additionally incorporates reflected signals. Li et al.~\cite{li3DUltrasoundSpine2021} reconstruct scoliotic spines from 3D ultrasound, separating soft tissue and bone better than voxel-based nearest-neighbour methods. Dynamic extensions condition the field on time: D-NeRF~\cite{pumarolaDNeRFNeuralRadiance2020} factorises a scene into a static canonical field and a learned time-dependent deformation, and EndoNeRF~\cite{wangNeuralRenderingStereo2022} applied this to binocular endoscopic video to remove instrument occlusions. Dynamic scene reconstruction from 2D ultrasound with neural fields remains largely unexplored.

\paragraph{3D Flow Reconstruction}

Neural fields have recently been used for 3D+t cardiovascular flow. Garzia et al.~\cite{garziaNeuralFieldsContinuous2025} model wall motion to constrain 4D Flow MRI reconstruction, and Wong et al. \cite{wong3DVelocityPressure2025a} reconstruct 3D velocity fields from 3D colour Doppler. Both, however, assume volumetric input and do not address sparse 2D measurements. For the latter, function-based interpolation such as radial basis functions (RBFs) has proven effective for reconstructing volumes from 2D slices \cite{huangReviewRealTime3D2017}. Poelma et al.~\cite{poelma3DFlowReconstruction2011} recover out-of-plane velocity from 2D acquisitions by combining EchoPIV in-plane fields with correlation peak strength, but rely on regularly spaced scans of a curved tube and a parabolic profile fit. RBFs can accurately approximate more complex profiles, motivating our investigation of whether continuous neural fields offer improved accuracy and flexibility for reconstructing transient 3D flow from sparse planar measurements.

\paragraph{Flow Reconstruction From Planes} Physics-informed networks have been used to infer 3D flow from planar velocity data. Cai et al.~\cite{caiPhysicsinformedNeuralNetworks2021} reconstruct a cylinder wake from a few axis-aligned planes, and Epshtein et al.~\cite{epshteinHighresolution3DFlow2026} use a projection loss for tilted planes in steady, low-Reynolds-number cerebrospinal fluid flow. Both depend on the Navier--Stokes residuals, and the latter also on wall boundary data, to recover the unobserved component. Both are evaluated on simulated data, and neither handles time-resolved flow on freely oriented planes. Triplane vector flow mapping~\cite{vixegeFullvolumeThreecomponentIntraventricular2022} measures one velocity component per plane and is specific to the left ventricle, while implicit representations for 4D Flow MRI~\cite{fathiSuperresolutionDenoising4DFlow2020, saittaImplicitNeuralRepresentations2024} require volumetric input. Our framework removes these restrictions: it reconstructs time-resolved flow from freely oriented two-component planes, requires neither governing equations nor boundary data, and is validated on three anatomies including real whole-heart measurements.

\section{Methodology}

\begin{figure*}[tb]
    \includegraphics[width=\linewidth]{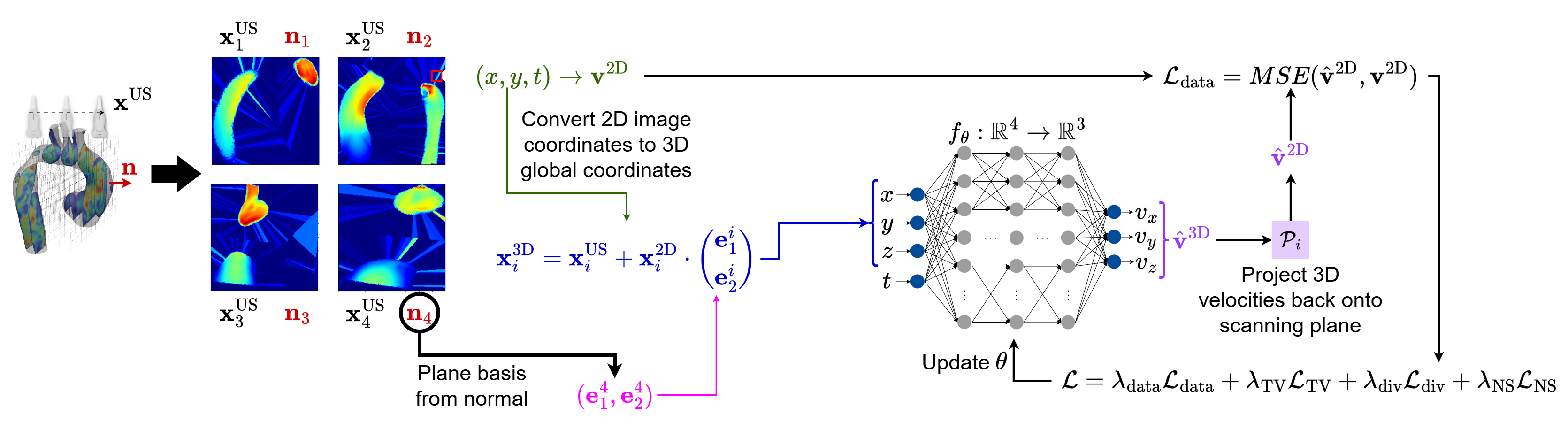}
    \caption{Our proposed framework takes 2D velocity fields, converts local coordinates into global 3D coordinates using the transducer position $\mathbf{x}_i^\text{US}$ and orientation $\mathbf{n}_i$, and a MLP predicts the 3D velocity vector. The loss function optimises the reconstruction of the 2D projection in accordance with physical constraints.}
    \label{fig:model}
\end{figure*}

Figure \ref{fig:model} shows a schematic of the proposed framework.

\subsection{Problem Formulation and Preprocessing}

Given a set of scans $\mathcal{D} = \{I_i \in \mathbb{R}^{T \times H \times W}\}_{i=1}^N$, velocimetry is performed to produce a 2D velocity field $U_i \in \mathbb{R}^{T \times H \times W \times 2}$ for each scan. Additionally, the position of the transducer $\mathbf{x}_i^\text{US}$ and the normal to the imaging plane $\mathbf{n}_i$ are recorded. The Cartesian grid produced by velocimetry exists in a local 2D image coordinate system, with coordinates $\mathbf{x}_i^\text{2D}$ for each scan $I_i$. To transform the points to a global coordinate system, a plane basis $(\mathbf{e}_1^i, \mathbf{e}_2^i)$ was formed from $\mathbf{n}_i$ and the transformation was defined as
\begin{equation}
    \mathbf{x}_i^\text{3D} = \mathbf{x}_i^\text{US} +
    \begin{pmatrix}
        \mathbf{e}_1^i \\
        \mathbf{e}_2^i
    \end{pmatrix}^\top \cdot\mathbf{x}_i^\text{2D} ,
\end{equation}
where $\mathbf{x}_i^\text{3D}$ is the point in the global 3D coordinate system. Points were masked based on manual segmentation of boundary walls, with points lying in the region of interest denoted by $\tilde{\mathbf{x}}_i$.

To render 2D views from the 3D flow field $\mathbf{u}: \mathbb{R}^3 \times \mathbb{R} \to \mathbb{R}^3$, the projection operator was defined as
\begin{equation}
    \mathcal{P}_j(\mathbf{v}) =
    \begin{pmatrix}
        \mathbf{e}_1^j \\
        \mathbf{e}_2^j
    \end{pmatrix} \cdot \mathbf{v},
\end{equation}
where $(\mathbf{e}_1^j, \mathbf{e}_2^j)$ are a plane basis from a known normal $\mathbf{n}_j$, and $\mathbf{v} \in \mathbb{R}^3$ is a sample from $\mathbf{u}$. The inverse problem consists of recovering $\mathbf{u}$ from a collection of sparse 2D observations acquired at different probe poses.

The models were evaluated with the relative root mean squared error (rRMSE) defined as
\begin{equation}
    \text{rRMSE} = \sqrt{\frac{\sum_{i=1}^N||v_i-v_i^*||^2}{\sum_{i=1}^N||v_i^*||^2}},
\end{equation}
where $v_i, v_i^* \in \mathbb{R}^3$ are the predicted and ground truth velocity vectors, respectively. The sums run over all evaluation points and time frames before the ratio is taken, so regions of low velocity are not weighted more heavily than regions of high velocity.

\subsection{Neural Field Model}

We use a SiLU activated neural network parameterised by $\theta$, $f_\theta: \mathbb{R}^4 \to \mathbb{R}^3$, to represent the 3D+t velocity field $\mathbf{u}$ with the transformed 3D coordinates, $\mathbf{x}_i^\text{3D}$, and time as inputs. The plain MLP uses SiLU activations, and the physics-constrained SIREN (PC-SIREN) uses sine activations~\cite{sitzmannImplicitNeuralRepresentations2020a}. The two-branch architecture used in D-NeRF~\cite{pumarolaDNeRFNeuralRadiance2020} was not an appropriate choice for our problem as it operates under the assumption that the transient velocity field can be explained by deforming a canonical field, which is not generally a physically natural assumption for fluid flow. The deformation network may also be susceptible to warping the field in ways that fit observations but produce physically implausible dynamics. Therefore, most suitable was a network expressing the velocity field as a single function of space and time, in conjunction with the physics-based loss function comprising of three parts:

\paragraph{Data Reconstruction Loss} The reconstruction loss was defined as
\begin{equation}
    \mathcal{L}_\text{data} = \frac{1}{N}\sum_{k=1}^N ||\mathcal{P}_{i(k)}[f_\theta(\tilde{\mathbf{x}}_k, t_k)] - \mathbf{u}^\text{2D}_k||_2^2,
\end{equation}
where $\mathbf{u}^\text{2D}_k \in \mathbb{R}^2$ is the in-plane velocity measured at sample $k$, $(\tilde{\mathbf{x}}_k, t_k)$ is its position and time, $i(k)$ is the scan it belongs to, and $||\cdot||_2$ is the $L_2$-norm.

\paragraph{Smoothness Loss} To mitigate the effect of noise in the 2D measurements, a first-order smoothness loss, denoted $\mathcal{L}_\text{TV}$, was defined as
\begin{equation}
    \mathcal{L}_\text{TV} = \frac{1}{M}\sum_{i=1}^M ||f_\theta(\tilde{\mathbf{x}}_i^\text{3D},t) - f_\theta(\tilde{\mathbf{x}}_i^\text{3D} + \delta \mathbf{x}, t)||_2^2,
\end{equation}
where $M$ is the number of samples and $\delta\mathbf{x}$ is a small perturbation in each spatial component.

\paragraph{Physics Loss} To ensure the reconstructed velocity field conserved mass and momentum, the divergence loss was defined as
\begin{equation}
    \mathcal{L}_\text{div} = \frac{1}{M}\sum_{i=1}^M |\nabla \cdot f_\theta(\tilde{\mathbf{x}}_i^\text{3D},t)|,
\end{equation}
and the velocity-vorticity formulation of the Navier-Stokes loss~\cite{jinNSFnetsNavierStokesFlow2021} was defined as
\begin{equation}
\begin{split}
    \mathcal{L}_\text{NS} = \frac{1}{M} \sum_{i=1}^M \left|\frac{\partial \omega}{\partial t} + \nabla \times (\omega \times f_\theta(\tilde{\mathbf{x}}_i^\text{3D},t)) \right . \\ 
    \left . + \frac{1}{\text{Re}} \nabla \times \nabla \times \omega\right|,
\end{split}
\end{equation}
where $\omega = \nabla \times f_\theta$ is the vorticity.

The total loss function was
\begin{equation}\label{eq:total_loss}
    \mathcal{L} = \lambda_\text{data}\mathcal{L}_\text{data} + \lambda_\text{TV}\mathcal{L}_\text{TV} + \lambda_\text{div}\mathcal{L}_\text{div} + \lambda_\text{NS}\mathcal{L}_\text{NS},
\end{equation}
where $\lambda_i \in \mathbb{R}$ are loss weights. The plain MLP uses $\mathcal{L}_\text{data}$ only, so it needs no governing equations or boundary conditions.

\subsection{Projection-Aware RBF Interpolation Baseline}
% description of how it works

Given a set of observed positions and their associated velocity measurements, RBF methods approximate the underlying field using distance-dependent basis functions at the observations~\cite{huangReviewRealTime3D2017}. In the standard formulation, a scalar field $f(\mathbf{x})$ is represented as 
\begin{equation}
    f(\mathbf{x}) = \sum_{i=1}^N w_i \phi(|\mathbf{x} - \mathbf{x}_i|),
\end{equation}
where $\mathbf{x}_i$ denotes the $i$-th measurement location, $\phi$ is a radial basis function, and $w_i$ are coefficients determined from the observed data, potentially with a smoothing term. For velocity reconstruction, however, measurements acquired on differently oriented planes provide components in different local coordinate systems. Consequently, these components can not be directly interpolated as independent global velocity components. Instead, at a query location $\mathbf{x}$, the 3D velocity estimate $\hat{\mathbf{v}}(\mathbf{x},t)$ is obtained by combining neighbouring observations according to their spatial proximity while accounting for each measurement plane's orientation. Let $\mathbf{P}_i \in \mathbb{R}^{2\times3}$ denote the projection matrix whose rows are the two in-plane unit basis vectors, and let $\mathbf{u} \in \mathbb{R}^2$ denote the measured in-plane velocity. The local estimate can then be formulated as a regularised weighted least-squares problem
\begin{equation}
    \hat{\mathbf{v}}(\mathbf{x},t) = \argmin_{\mathbf{v} \in \mathbb{R}^3} \sum_{i \in \mathcal{N}(\mathbf{x},t)} w_i(\mathbf{x},t) ||\mathbf{P}_i\mathbf{v} - \mathbf{u}_i||_2^2 + \lambda||\mathbf{v}||_2^2,
\end{equation}
where $\mathcal{N}(\mathbf{x},t)$ denotes the selected neighbouring measurements, $w_i$ are distance-dependent weights, and $\lambda$ controls regularisation. This formulation accounts for the orientations of the measurement planes when estimating the global velocity vector. The resulting estimate is evaluated at query locations throughout the reconstruction domain to obtain a continuous approximation of the flow field.
\section{Experimental Setup}

\subsection{Datasets}

\paragraph{Coarctation of Aorta (CoA)} The AortaPINNs dataset~\cite{khanAortaPINNsData2025} contained computational fluid dynamics (CFD) simulations of an aorta geometry at different levels of coarctation, quantified by the relative diameter of the narrowed region. In this work, simulations with 30\%, 40\%, 60\%, 80\%, and 100\% of the full diameter were used to capture a broad variation in haemodynamics.

\paragraph{Intracranial Aneurysms (IA)} The AnXplore dataset~\cite{goetzAnXploreComprehensiveFluidstructure2024} contained fluid-structure interaction (FSI) simulations of various aneurysm geometries obtained from real patient cases with the vessel geometry kept constant to ensure that variation in dynamics were caused by aneurysm sac geometry alone. In this work, volume of the aneurysm sac was varied as it has been shown to be a strong indicator of risk of rupture and significantly alters blood flow~\cite{dharMorphologyParametersIntracranial2008b}.

\paragraph{Full Heart MRI} Arshad et al.~\cite{arshadMotionrobustFreerunningVolumetric2024} performed a thoracic 4D Flow MRI scan of a patient at rest and manual segmentation of the main vessels and chambers was performed with 3D Slicer (v5.12.4). Velocity vectors were masked according to the result of the segmentation.

\subsection{Implementation}

Each instance of the model was trained using two NVIDIA RTX A5000 GPUs with 24GB memory and a hold-out scan was removed from the training set for evaluation of novel-view synthesis. For the CoA and IA datasets, the sliced 2D flow fields were $200\times200$ pixels, but the slices of the MRI dataset were $200\times400$ pixels. Each instance of the model was evaluated on 30000 points in the CoA dataset, 10000 points in the IA dataset, and 38985 points in the MRI dataset. The values in Tables \ref{tab:aorta}, \ref{tab:aneurysm} and \ref{tab:aorta_res} are means over three training runs with different random seeds. Implementation details, including the training settings, are given in the supplementary material, and we defer further results to the supplementary material.

\subsection{Scanning Configurations}

\begin{figure}[h]
    \centering
    \includegraphics[width=\linewidth]{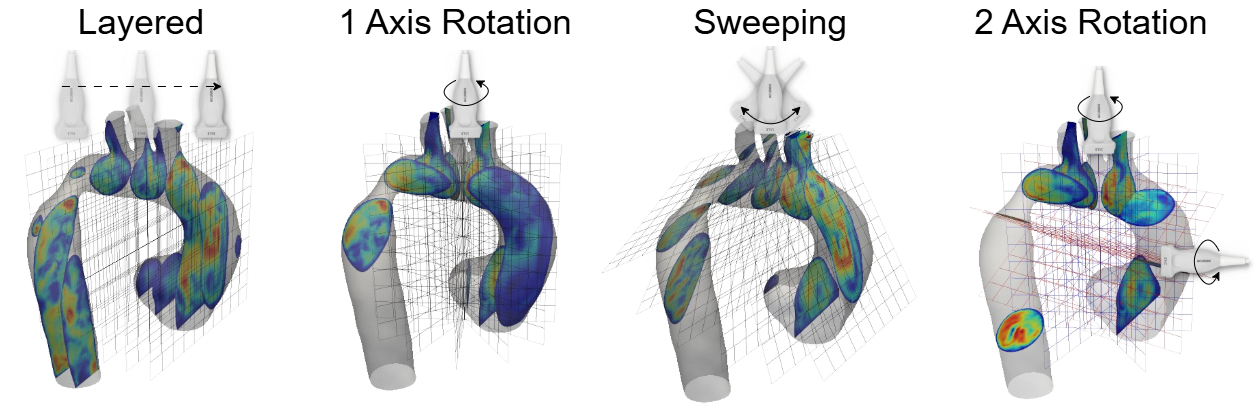}
    \caption{Visualisation of the different scanning modes that were tested.}
    \label{fig:scanning_modes}
\end{figure}

Figure \ref{fig:scanning_modes} shows the different ways that scans could be acquired by either translating the transducer along an axis or by rotating it about various axes. For each scanning mode, training data was generated with every combination of axes for a certain geometry (CoA with $60\%$ coarctation) and a plain multi-layer perceptron (MLP) was trained for each scanning mode and axis. The best configuration was used for the remaining experiments.

\subsection{Ablations}

Experiments initially used a plain MLP, trained using only $\mathcal{L}_\text{data}$, without coordinate encoding or physics-based regularisation. First, the number of scans was fixed at 20 to compare different scanning configurations and identify the most effective acquisition strategy. The selected configuration was then fixed while the number of scans was varied between 4 and 100. The final scan count was chosen as a compromise between reconstruction accuracy and training time. The same procedure was repeated for the classical RBF baseline on the same scans, independently selecting its optimal scanning configuration and number of scans to ensure a fair comparison.

With the scan configuration and number of scans fixed, different coordinate representations were investigated to assess the contribution of encoding to the representation of complex flow fields. These included using only scaled spatial and temporal coordinates without additional encoding, Fourier feature encoding~\cite{tancikFourierFeaturesLet2020} (applied to spatial and temporal coordinates separately and jointly), and a SIREN~\cite{sitzmannImplicitNeuralRepresentations2020a} network. Finally, the loss weights in Eq. (\ref{eq:total_loss}) were optimised. A sequential approach was adopted in which individual loss weights were varied while the remaining weights were held fixed. The reconstruction weight $\lambda_\text{data}$ was maintained at 1 throughout. First, $\lambda_\text{NS}$ was varied with $\lambda_\text{div}=\lambda_\text{TV}=0$. Next, the selected $\lambda_\text{NS}$ was fixed while $\lambda_\text{div}$ was varied. Finally, the selected $\lambda_\text{NS}$ and $\lambda_\text{div}$ were fixed while $\lambda_\text{TV}$ was varied. The order was chosen based on the complexity of the individual loss function, and the procedure was used to determine the final loss weighting for the proposed model. All of the selection described above was carried out on the 60\% coarctation case, and the selected configuration was applied unchanged to all other cases.
\section{Results and Discussion}

\subsection{Acquisition Strategy and Model Selection}
The two-axis rotational scanning configuration achieved the best 3D reconstruction accuracy for both the MLP-based and RBF models. In particular, sampling across two rotational axes provided more diverse measurement orientations, enabling better reconstruction of out-of-plane velocity components compared with the layered scanning configuration. Reconstruction accuracy improved rapidly as the number of scans increased, with performance largely converging at approximately 50 scans. However, the improvement between 20 and 50 scans was marginal, despite the latter requiring over twice the training time. Therefore, 20 scans were selected as a compromise between reconstruction accuracy and computational cost.

For the aorta dataset, SIREN encoding achieved substantially better reconstruction accuracy than the Fourier feature encoding strategies and a modest improvement over the plain MLP. Consequently, SIREN was selected for the proposed model. The final loss weights were determined using the sequential optimisation procedure described above, yielding $\lambda_\text{data} = 1$, $\lambda_\text{TV} = 10^{-3}$, $\lambda_\text{div} = 10^{-2}$, and $\lambda_\text{NS} = 10^{-3}$. This physics-constrained SIREN model is denoted as \emph{PC-SIREN}.

\subsection{Reconstruction Performance}
\begin{figure*}[tb]
    \includegraphics[width=\linewidth]{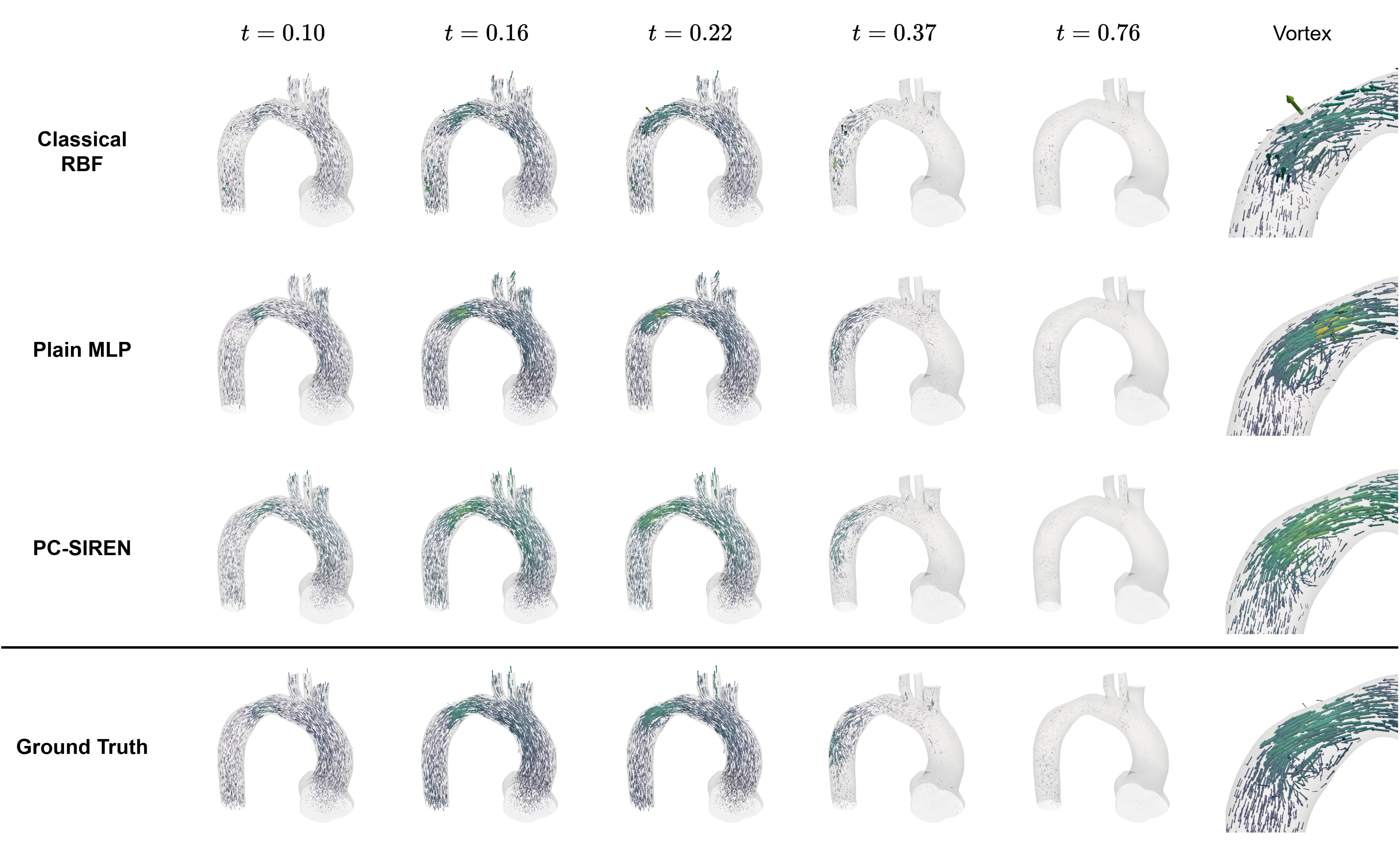}
    \caption{3D Reconstruction of the aorta geometry with 60\% radius. Though similar, PC-SIREN captures the transition between the vortex and downstream flow smoother than the plain MLP.}
    \label{fig:aorta}
\end{figure*}

Table \ref{tab:aorta} shows that the PC-SIREN model achieves lowest rRMSE for both the full 3D volume and the held-out measurements when reconstructing the aorta with 60\% coarctation severity, outperforming both the plain MLP and the RBF baseline. This quantitative improvement is also reflected in the reconstructions shown in Figure \ref{fig:aorta} across multiple timesteps. While the plain MLP captures the overall flow structure, its reconstruction exhibits slightly greater spatial variation, whereas PC-SIREN produces a smoother and more coherent velocity field while retaining prominent flow features, including the vortex. In contrast, the RBF baseline provides a less continuous representation of the underlying flow.

\begin{table}[h]
    \centering
    \caption{rRMSE of predicted 3D velocity field and hold-out scan for aorta geometry with 60\% radius.}
    \begin{tabular}{l c c}
        \toprule
        \textbf{Model} & \textbf{3D rRMSE} & \textbf{Hold-out rRMSE} \\
        \midrule
        RBF & 0.7256 & 0.7696 \\
        Plain MLP & 0.4273 & 0.7623 \\
        PC-SIREN & $\mathbf{0.3756}$ & $\mathbf{0.4950}$ \\
        \bottomrule
    \end{tabular}
    \label{tab:aorta}
\end{table}

\begin{figure}[h]
    \centering
    \includegraphics[width=\linewidth]{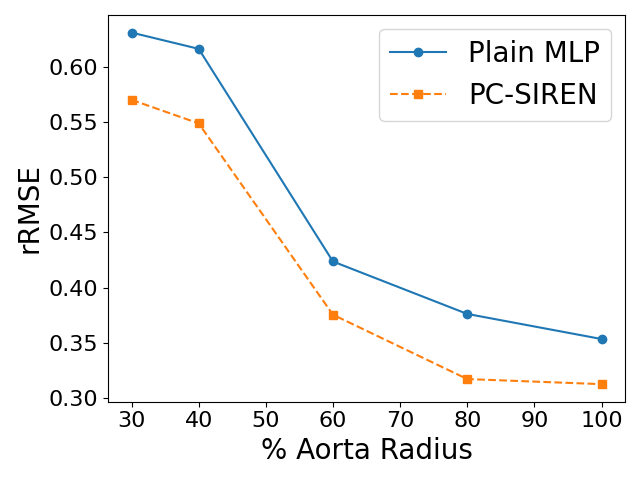}
    \caption{PC-SIREN produced more accurate reconstructions than the plain MLP at all coarctation levels. Both models saw improved reconstruction quality as the geometry approached a healthier regime.}
    \label{fig:coa_level}
\end{figure}

Figure \ref{fig:coa_level} shows that the reconstruction quality improves as the level of coarctation approaches a healthier regime (100\% is a healthy aorta), suggesting that smoother flows with fewer disturbances are less challenging for the reconstruction framework. Importantly, the PC-SIREN model consistently achieves a lower rRMSE than the plain MLP at all coarctation levels.

For the aneurysms, the plain MLP is the most accurate model on all five samples, with an rRMSE between 0.07 and 0.16, a reduction of 72 to 89\% relative to the RBF baseline (Table \ref{tab:aneurysm}). The framework therefore reconstructs these flows accurately from the projection loss alone, with no governing equations and no tuning beyond the aorta. PC-SIREN, whose configuration was selected on the 60\% coarctation case and applied unchanged, does not improve on the plain MLP here. Furthermore, the aneurysm dataset exhibits relatively organised and less spatially varying flow patterns, which may be adequately represented by the lower-frequency functions learned by a plain MLP. In comparison, the higher-frequency representational capacity of SIREN may be unnecessary for these smoother flow fields and could make optimisation more challenging. Because the network is interchangeable, the representation can be chosen per flow type without changing the framework.
\begin{table}[h]
    \centering
    \caption{rRMSE of 3D reconstruction of aneurysm flow.}
    \begin{tabular}{l c c c}
        \toprule
        \textbf{Sample} & \textbf{RBF} & \textbf{Plain MLP} & \textbf{PC-SIREN} \\
        \midrule
        \#11 & 0.5565 & \textbf{0.1183} & 0.5662 \\
        \#117 & 0.5867 & \textbf{0.1635} & 0.5543 \\
        \#161 & 0.5724 & \textbf{0.1222} & 0.6640 \\
        \#183 & 0.6196 & \textbf{0.0706} & 0.6171 \\
        \#195 & 0.5719 & \textbf{0.0773} & 0.7141 \\
        \bottomrule
    \end{tabular}
    \label{tab:aneurysm}
\end{table}

\subsection{Sensitivity of Resolution and Noise}

%The contrasting performance across the aorta and aneurysm datasets suggests that the relative benefits of PC-SIREN and a plain MLP depend on the characteristics of the underlying flow. To investigate this further, reconstruction accuracy was evaluated across different spatial resolutions for the aorta with 60\% coarctation severity and the aneurysm dataset. As shown in Table \ref{tab:aorta_res}, PC-SIREN consistently outperforms the plain MLP across all tested resolutions for the aorta, whereas the opposite trend is observed for the aneurysm samples (Figure \ref{fig:resolution}). Interestingly, increasing spatial resolution reduces reconstruction error for the aneurysm, although improvements begin to plateau at $100\times100$px. In contrast, the aorta achieves lower reconstruction error at $100\times100$ than at $200\times200$, indicating that denser sampling does not necessarily improve reconstruction accuaracy. These differences may reflect the distinct flow characteristics of the datasets: the more spatially complex aortic flow may benefit from SIREN's higher-frequency representational capacity and physics regularisation, whereas the relatively smooth and organised aneurysm flow may be adequately represented by a plain MLP. Overall, these findings suggest that both network architecture and measurement resolution should be considered in relation to the underlying flow characteristics.

\begin{figure}[h]
    \centering
    \includegraphics[width=\linewidth]{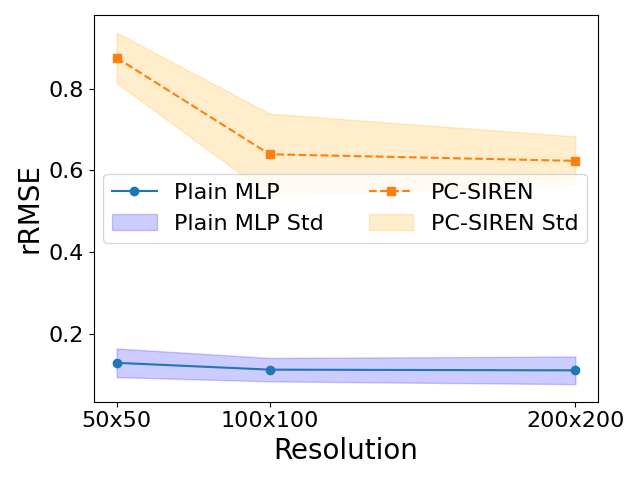}
    \caption{Effect of scan resolution on rRMSE for the aneurysm data. The plain MLP is accurate at all resolutions. Table \ref{tab:aorta_res} gives the corresponding results for the aorta.}
    \label{fig:resolution}
\end{figure}

The contrasting results across the aorta and aneurysm datasets suggest that model performance depends on the underlying flow characteristics. Across all tested resolutions, PC-SIREN outperforms the plain MLP for the aorta, whereas the opposite is observed for the aneurysm (Table \ref{tab:aorta_res}, Figure \ref{fig:resolution}). Increasing resolution improves aneurysm reconstruction, although gains plateau at $100\times100$px. For the aorta, PC-SIREN is robust to coarse sampling: at $50\times50$px, with sixteen times fewer samples per scan, its rRMSE of 0.39 is still lower than that of the plain MLP (0.43) and RBF (0.73) at $200\times200$px. The contrast between the two datasets may reflect differences in flow complexity: the more spatially varying aortic flow may benefit from SIREN's higher-frequency representational capacity and physics-based regularisation, while the smoother aneurysm flow may be better suited to a plain MLP. These findings highlight the importance of considering flow characteristics when selecting both network architecture and measurement resolution.

\begin{table}[h]
    \centering
    \caption{Effect of scan resolution on rRMSE for aorta with 60\% radius.}
    \begin{tabular}{l c c c}
    \toprule
        \textbf{Model} & $\mathbf{50\times50}$ & $\mathbf{100\times100}$ & $\mathbf{200\times200}$ \\
        \midrule
        Plain MLP & 0.5231 & 0.4387 & 0.4273 \\
        PC-SIREN & \textbf{0.3898} & \textbf{0.3659} & \textbf{0.3756} \\
        \bottomrule
    \end{tabular}
    \label{tab:aorta_res}
\end{table}

%However, the plain MLP consistently and substantially outperforms the proposed method across all tested resolutions with less variation. This suggests that, under the spatial sampling conditions considered, the additional representational capacity and physics-based constraints of the proposed model do not necessarily provide an advantage. Instead, the simpler MLP appears better suited to reconstructing the relatively organised flow structures in this aneurysm sample, particularly when the available measurements are spatially limited.

%The effect of multiplicative Rayleigh noise on reconstruction performance was investigated at three noise scales ($1.0$, $5.0$, and $10.0$) for the aorta and aneurysm datasets. Table \ref{tab:noise} shows that the RBF baseline consistently achieves the poorest performance, exhibiting negative improvement relative to the noisy input at all noise levels for both datasets. For the aneurysm, the plain MLP provides positive improvement at lower noise levels but deteriorates as noise increases, while PC-SIREN exhibits negative improvement at all tested levels. Nevertheless, PC-SIREN becomes more competitive with the MLP at higher noise levels. In contrast, for the aorta, PC-SIREN achieves positive improvement at the lowest noise level and consistently outperforms both alternative methods across all tested noise levels., although its improvement over the noisy input becomes negative as noise increases. This difference is consistent with the earlier results: physics-based regularisation provides a clear benefit for more complex aortic flow, potentially helping to suppress measurement noise. However, since it did not improve reconstruction accuracy for the aneurysm dataset in the absence of added noise, its limited ability to improve noisy aneurysm reconstructions is perhaps unsurprising. Overall, these findings suggest that the benefits of PC-SIREN extend beyond reconstruction accuracy to the handling of noisy measurements, but remain dependent on the flow being reconstructed.

The effect of multiplicative Rayleigh noise was evaluated at three scales ($1.0$, $5.0$, and $10.0$) for both datasets (Table \ref{tab:noise}). Improvement is the reduction in rRMSE relative to the noisy input. At scales 5.0 and 10.0 the noise is several times larger than the signal, so these rows are a stress test.RBF shows negative improvement at all noise levels. For the aneurysm, the plain MLP improves on the noisy input at low noise but deteriorates at higher levels, while PC-SIREN remains below the noisy input but becomes more competitive with the MLP as noise increases. In contrast, PC-SIREN outperforms both alternatives for the aorta, except at the highest level of noise where RBF achieves marginally better performance. Nevertheless, PC-SIREN consistently outperforms the plain MLP for the aorta across all tested noise levels. This supports the earlier findings that physics-based regularisation benefits the more complex aortic flow and may help mitigate measurement noise. Its limited benefit for aneurysm reconstruction, however, is also reflected in its response to noise. Overall, the effectiveness of PC-SIREN in handling noisy measurements appears dependent on the underlying flow characteristics and noise level.

% \begin{table}[h]
%     \centering
%     \caption{Effect of Rayleigh noise on rRMSE for aneurysm \#11}
%     \scalebox{0.9}{
%     \begin{tabular}{l c c c c c}
%     \toprule
%         \textbf{Scale} & \textbf{Input} & \textbf{Plain MLP} & \textbf{\% Imp.} & \textbf{Ours} & \textbf{\% Imp.} \\
%         \midrule
%         $1.0$ & 0.4735 & 0.2441 & \textbf{48.45} & 0.5788 & $-22.24$ \\
%         $5.0$ & 4.1830 & 5.0925 & $-21.74$ & 4.3008 & $\mathbf{-2.816}$ \\
%         $10.0$ & 8.9440 & 11.1669 & $-24.85$ & 9.3828 & $\mathbf{-4.906}$ \\
%         \bottomrule
%     \end{tabular}}
%     \label{tab:noise}
% \end{table}

\begin{table}[h]
    \centering
    \caption{Effect of Rayleigh noise on rRMSE for aneurysm \#11 (top) and aorta with 60\% radius (bottom).}
    \scalebox{0.8}{
    \begin{tabular}{l c c c c}
    \toprule
         \multicolumn{1}{c}{\multirow{2}{*}{\textbf{Scale}}} & \multicolumn{1}{c}{\multirow{2}{*}{\textbf{Input rRMSE}}} & \multicolumn{3}{c}{\textbf{\% Improvement}} \\
        \multicolumn{1}{c}{} & \multicolumn{1}{c}{} & \textbf{RBF} & \textbf{MLP} & \textbf{PC-SIREN} \\
        \midrule
        $1.0$ & 0.4735 & $-37.1196$ & \textbf{48.4460} & $-22.1539$ \\
        $5.0$ & 4.1830 & $-26.0619$ & $-21.7422$ & $\mathbf{-2.8498}$ \\
        $10.0$ & 8.9440 & $-25.9106$ & $-24.8533$ & $\mathbf{-4.8789}$ \\
        \midrule
        $1.0$ & 0.4195 & $-91.5792$ & $-59.5948$ & \textbf{5.0322} \\
        $5.0$ & 3.7051 & $-25.1175$ & $-35.6834$ & $\mathbf{-22.6175}$ \\
        $10.0$ & 7.9241 & $\mathbf{-24.4633}$ & $-31.4638$ & $-25.9701$ \\
        \bottomrule
    \end{tabular}}
    \label{tab:noise}
\end{table}

\begin{figure}[t]
    \centering
    \includegraphics[width=\linewidth]{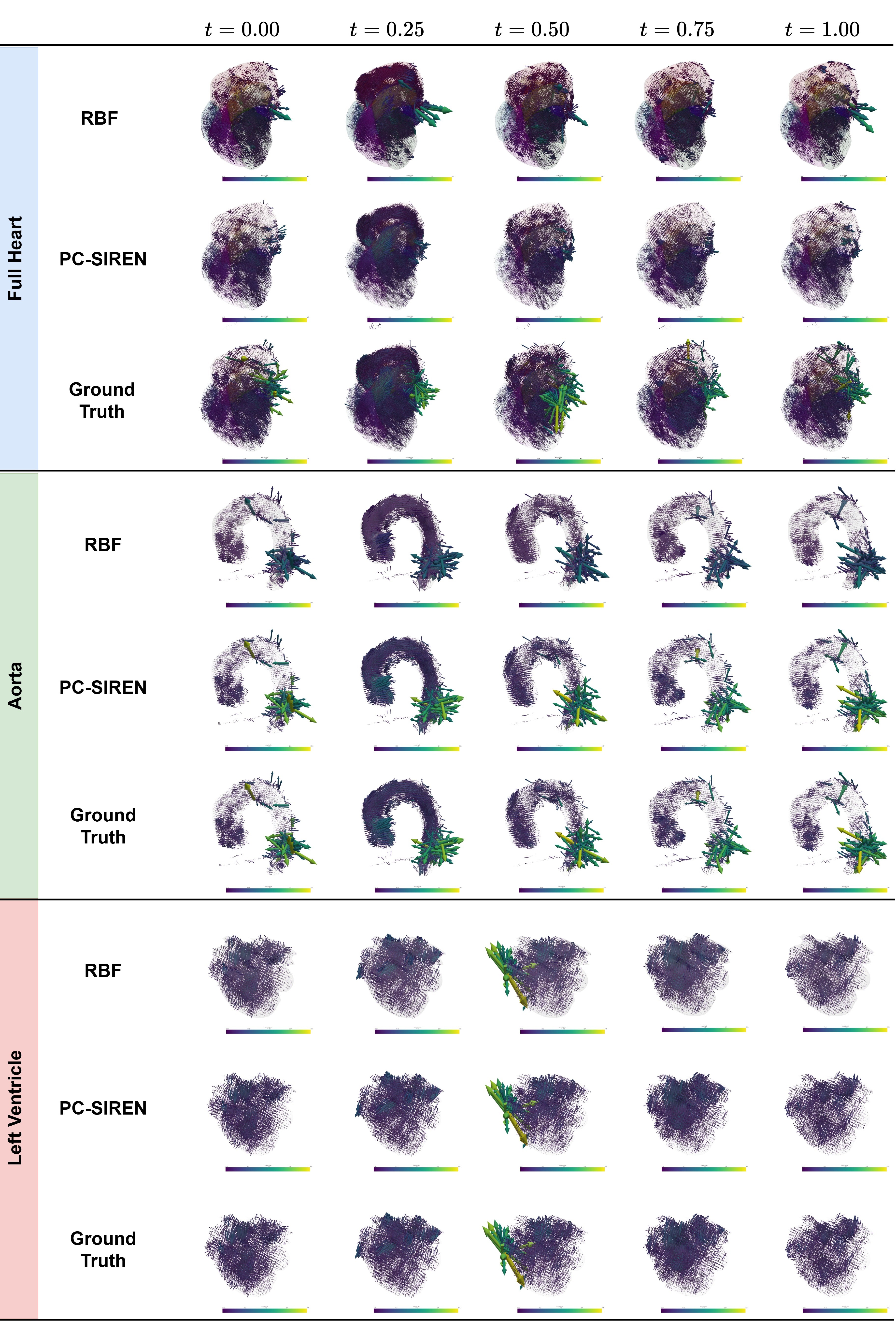}
    \caption{Qualitative comparison of RBF baseline and PC-SIREN model with ground truth 4D Flow MRI measurements of the full heart, aorta, and left ventricle.}
    \label{fig:mri}
\end{figure}

\subsection{Generalisation to Real Cardiac Data}

For the real cardiac MRI dataset, measurement planes were positioned using subcostal and suprasternal~\cite{henryReportAmericanSociety1980} transducer locations (below and above the sternum, respectively), as these provided convenient access for establishing two-axis rotational scanning with approximately perpendicular rotation axes. The in-plane sampling resolution was also changed from square to rectangular to accommodate the greater depth of the cardiac region of interest. Finally, measurements were masked using the overall region of interest rather than individual anatomical segments. This ensured that the reconstructed domain remained a single connected region, avoiding the introduction of multiple disconnected subdomains that could otherwise be treated as separate closed systems by the reconstruction framework.

Table \ref{tab:mri} shows that the PC-SIREN model achieves lower rRMSE than both the RBF baseline and plain MLP across the full heart and all evaluated anatomical segments, indicating improved reconstruction accuracy. The accuracy is also reflected in the qualitative comparisons in Figure \ref{fig:mri}, where PC-SIREN closely reproduces the ground-truth velocity fields across the full heart, aorta, and left ventricle. However, the higher-frequency representational capacity of SIREN may also allow it to fit local variations arising from measurement noise in the original 4D Flow MRI data. In contrast, the plain MLP's bias towards lower-frequency representations can suppress some of these noisy variations, producing smoother reconstructions, albeit with slightly reduced agreement with the measured ground truth.

\begin{table}[h]
    \centering
    \caption{Quantitative comparison of models (rRMSE) on the real cardiac 4D FLow MRI dataset.}
    \scalebox{0.95}{
    \begin{tabular}{l c c c}
        \toprule
        \textbf{Segment} & \textbf{RBF} & \textbf{Plain MLP} & \textbf{PC-SIREN} \\
        \midrule
        Full Heart & 0.7687 & 0.7184 & \textbf{0.6101} \\
        Aorta & 0.7827 & 0.7259 & \textbf{0.6303} \\
        Left Atrium & 0.7081 & 0.7248 & \textbf{0.6195} \\
        Left Ventricle & 0.8606 & 0.7700 & \textbf{0.6340} \\
        Pulmonary Artery & 0.6699 & 0.6185 & \textbf{0.5234} \\
        Right Atrium & 0.7117 & 0.6786 & \textbf{0.5920} \\
        Right Ventricle & 0.8172 & 0.7141 & \textbf{0.5729} \\
        \bottomrule
    \end{tabular}}
    \label{tab:mri}
\end{table}

As the reference 4D Flow MRI data contain measurement noise, part of the remaining error reflects noise in the reference rather than in the reconstruction. From scans acquired through only two acoustic windows, the framework reduces the rRMSE of RBF interpolation in every anatomical segment, by 13 to 30\%. Overall, these results demonstrate that the proposed framework can reconstruct cardiovascular velocity fields to a level of agreement comparable to 4D Flow MRI measurements.
\section{Conclusion}
We presented a plane-aware neural-field framework that reconstructs time-resolved 3D cardiovascular flow from sparse, freely oriented planar velocity measurements, without requiring governing equations, boundary conditions, or pressure data. It reduced the rRMSE of RBF interpolation by 48\% for aortic coarctation, 72 to 89\% for intracranial aneurysms, and 13 to 30\% across the whole heart in real 4D Flow MRI. We investigated different acquisition strategies, sampling densities, and network architectures. A plain MLP was sufficient for the aneurysms, while a physics-constrained SIREN added accuracy at each coarctation level and in every cardiac segment, and remained accurate with sixteen times fewer samples per scan. Under increasing noise, PC-SIREN generally became more competitive, although RBF remained marginally better at the highest noise level for the aorta. Future work will address probe pose estimation, ultrasound-derived velocities, and a larger cohort.

\section*{LLM Usage Statement}
LLM usage (Claude) was limited to editing and refining author-produced content. We can confirm that neither the full manuscript, or the major parts of it, were given to an LLM for reviewer-style feedback. This conception of this work was entirely author-produced.
{
    \small
    \bibliographystyle{ieeenat_fullname}
    \bibliography{3d_rendering}

@article{arshadMotionrobustFreerunningVolumetric2024,
  title = {Motion-Robust Free-Running Volumetric Cardiovascular {{MRI}}},
  author = {Arshad, Syed M. and Potter, Lee C. and Chen, Chong and Liu, Yingmin and Chandrasekaran, Preethi and Crabtree, Christopher and Tong, Matthew S. and Simonetti, Orlando P. and Han, Yuchi and Ahmad, Rizwan},
  year = 2024,
  month = sep,
  journal = {Magnetic Resonance in Medicine},
  volume = {92},
  number = {3},
  pages = {1248--1262},
  publisher = {John Wiley \& Sons, Ltd},
  issn = {1522-2594},
  doi = {10.1002/mrm.30123},
  urldate = {2026-10-06},
  langid = {english}
}

@article{avesaniVectorFlowMapping2021,
  title = {Vector Flow Mapping: {{A}} Review from Theory to Practice},
  shorttitle = {Vector Flow Mapping},
  author = {Avesani, Martina and Degrelle, Bastien and Di Salvo, Giovanni and Thambo, Jean-Benoit and Iriart, Xavier},
  year = 2021,
  month = aug,
  journal = {Echocardiography},
  volume = {38},
  number = {8},
  pages = {1405--1413},
  address = {Mount Kisco, N.Y.},
  issn = {1540-8175},
  doi = {10.1111/echo.15154},
  langid = {english},
  pmid = {34259359}
}

@article{chaouiThreedimensional3D4D2004,
  title = {Three-Dimensional ({{3D}}) and {{4D}} Color {{Doppler}} Fetal Echocardiography Using Spatio-Temporal Image Correlation ({{STIC}})},
  author = {Chaoui, R. and Hoffmann, J. and Heling, K. S.},
  year = 2004,
  journal = {Ultrasound in Obstetrics \& Gynecology},
  volume = {23},
  number = {6},
  pages = {535--545},
  issn = {1469-0705},
  doi = {10.1002/uog.1075},
  urldate = {2026-10-08},
  langid = {english}
}

@article{chongGlobalBurdenCardiovascular2025,
  title = {Global Burden of Cardiovascular Diseases: Projections from 2025 to 2050},
  shorttitle = {Global Burden of Cardiovascular Diseases},
  author = {Chong, Bryan and Jayabaskaran, Jayanth and Jauhari, Silingga Metta and Chan, Siew Pang and Goh, Rachel and Kueh, Martin Tze Wah and Li, Henry and Chin, Yip Han and Kong, Gwyneth and Anand, Vickram Vijay and Wang, Jiong-Wei and Muthiah, Mark and Jain, Vardhmaan and Mehta, Anurag and Lim, Shir Lynn and Foo, Roger and Figtree, Gemma A. and Nicholls, Stephen J. and Mamas, Mamas A. and Januzzi, James L. and Chew, Nicholas W. S. and Richards, A. Mark and Chan, Mark Y.},
  year = 2025,
  month = aug,
  journal = {European Journal of Preventive Cardiology},
  volume = {32},
  number = {11},
  pages = {1001--1015},
  publisher = {Oxford Academic},
  issn = {2047-4873},
  doi = {10.1093/eurjpc/zwae281},
  urldate = {2026-01-15},
  langid = {english}
}

@inproceedings{corona-figueroaMedNeRFMedicalNeural2022,
  title = {{{MedNeRF}}: {{Medical Neural Radiance Fields}} for {{Reconstructing 3D}}-Aware {{CT-Projections}} from a {{Single X}}-Ray},
  shorttitle = {{{MedNeRF}}},
  booktitle = {2022 44th {{Annual International Conference}} of the {{IEEE Engineering}} in {{Medicine}} \& {{Biology Society}} ({{EMBC}})},
  author = {{Corona-Figueroa}, Abril and Frawley, Jonathan and Taylor, Sam Bond- and Bethapudi, Sarath and Shum, Hubert P. H. and Willcocks, Chris G.},
  year = 2022,
  month = jul,
  pages = {3843--3848},
  issn = {2694-0604},
  doi = {10.1109/EMBC48229.2022.9871757},
  urldate = {2026-10-26}
}

@article{dharMorphologyParametersIntracranial2008b,
  title = {Morphology {{Parameters}} for {{Intracranial Aneurysm Rupture Risk Assessment}}},
  author = {Dhar, Sujan and Tremmel, Markus and Mocco, J and Kim, Minsuok and Yamamoto, Junichi and Siddiqui, Adnan H. and Hopkins, L. Nelson and Meng, Hui},
  year = 2008,
  month = aug,
  journal = {Neurosurgery},
  volume = {63},
  number = {2},
  pages = {185--197},
  issn = {0148-396X},
  doi = {10.1227/01.NEU.0000316847.64140.81},
  urldate = {2026-07-13},
  pmcid = {PMC2570753},
  pmid = {18797347}
}

@article{fensterThreeDimensionalUltrasoundImaging2000,
  title = {Three-{{Dimensional Ultrasound Imaging}}},
  author = {Fenster, Aaron and Downey, Donal B.},
  year = 2000,
  journal = {Annual Review of Biomedical Engineering},
  volume = {2},
  pages = {457--475},
  publisher = {Annual Reviews},
  issn = {1545-4274},
  doi = {10.1146/annurev.bioeng.2.1.457},
  urldate = {2026-10-08}
}

@article{forsbergClinicalApplicationsUltrasound1998,
  title = {Clinical Applications of Ultrasound Contrast Agents},
  author = {Forsberg, F and Merton, {\relax DA} and Liu, {\relax JB} and Needleman, L and Goldberg, {\relax BB}},
  year = 1998,
  journal = {Ultrasonics},
  volume = {36},
  number = {1-5},
  pages = {695--701},
  publisher = {Elsevier}
}

@inproceedings{garziaNeuralFieldsContinuous2025,
  title = {Neural {{Fields}} for~{{Continuous Periodic Motion Estimation}} in~{{4D Cardiovascular Imaging}}},
  booktitle = {Statistical {{Atlases}} and {{Computational Models}} of the {{Heart}}. {{Workshop}}, {{CMRxRecon}} and {{MBAS Challenge Papers}}.},
  author = {Garzia, Simone and Rygiel, Patryk and Dummer, Sven and Cademartiri, Filippo and Celi, Simona and Wolterink, Jelmer M.},
  editor = {Camara, Oscar and {Puyol-Ant{\'o}n}, Esther and Sermesant, Maxime and Suinesiaputra, Avan and Zhao, Jichao and Wang, Chengyan and Tao, Qian and Young, Alistair},
  year = 2025,
  pages = {378--389},
  publisher = {Springer Nature Switzerland},
  address = {Cham},
  doi = {10.1007/978-3-031-87756-8_37},
  isbn = {978-3-031-87756-8},
  langid = {english}
}

@article{goetzAnXploreComprehensiveFluidstructure2024,
  title = {{{AnXplore}}: A Comprehensive Fluid-Structure Interaction Study of 101 Intracranial Aneurysms},
  shorttitle = {{{AnXplore}}},
  author = {Goetz, Aur{\`e}le and {Jeken-Rico}, Pablo and Pelissier, Ugo and Chau, Yves and S{\'e}dat, Jacques and Hachem, Elie},
  year = 2024,
  month = jun,
  journal = {Frontiers in Bioengineering and Biotechnology},
  volume = {12},
  publisher = {Frontiers},
  issn = {2296-4185},
  doi = {10.3389/fbioe.2024.1433811},
  urldate = {2026-06-23},
  langid = {english}
}

@misc{guoUlReNeRF3DUltrasound2024,
  title = {{{UlRe-NeRF}}: {{3D Ultrasound Imaging}} through {{Neural Rendering}} with {{Ultrasound Reflection Direction Parameterization}}},
  shorttitle = {{{UlRe-NeRF}}},
  author = {Guo, Ziwen and Fang, Zi and Fu, Zhuang},
  year = 2024,
  month = sep,
  number = {arXiv:2408.00860},
  eprint = {2408.00860},
  primaryclass = {cs.AI},
  publisher = {arXiv},
  doi = {10.48550/arXiv.2408.00860},
  urldate = {2026-10-26},
  archiveprefix = {arXiv}
}

@article{henryReportAmericanSociety1980,
  title = {Report of the {{American Society}} of {{Echocardiography Committee}} on {{Nomenclature}} and {{Standards}} in {{Two}}-Dimensional {{Echocardiography}}.},
  author = {Henry, W L and DeMaria, A and Gramiak, R and King, D L and Kisslo, J A and Popp, R L and Sahn, D J and Schiller, N B and Tajik, A and Teichholz, L E and Weyman, A E},
  year = 1980,
  month = aug,
  journal = {Circulation},
  volume = {62},
  number = {2},
  pages = {212--217},
  publisher = {American Heart Association},
  doi = {10.1161/01.CIR.62.2.212},
  urldate = {2026-11-11}
}

@article{huangReviewRealTime3D2017,
  title = {A {{Review}} on {{Real-Time 3D Ultrasound Imaging Technology}}},
  author = {Huang, Qinghua and Zeng, Zhaozheng},
  year = 2017,
  month = jan,
  journal = {BioMed Research International},
  volume = {2017},
  number = {1},
  pages = {6027029},
  publisher = {John Wiley \& Sons, Ltd},
  issn = {2314-6141},
  doi = {10.1155/2017/6027029},
  urldate = {2026-10-06},
  langid = {english}
}

@article{jinNSFnetsNavierStokesFlow2021,
  title = {{{NSFnets}} ({{Navier-Stokes}} Flow Nets): {{Physics}}-Informed Neural Networks for the Incompressible {{Navier-Stokes}} Equations},
  author = {Jin, Xiaowei and Cai, Shengze and Li, Hui and Karniadakis, George Em},
  year = 2021,
  journal = {Journal of Computational Physics},
  volume = {426},
  pages = {109951},
  publisher = {Elsevier}
}

@misc{khanAortaPINNsData2025,
  type = {{{CFD}}},
  title = {{{AortaPINNsData}}},
  author = {Khan, Owais},
  year = 2025,
  month = jan,
  urldate = {2025-10-02}
}

@article{langEAEASERecommendations2012,
  title = {{{EAE}}/{{ASE Recommendations}} for {{Image Acquisition}} and {{Display Using Three-Dimensional Echocardiography}}},
  author = {Lang, Roberto M. and Badano, Luigi P. and Tsang, Wendy and Adams, David H. and Agricola, Eustachio and Buck, Thomas and Faletra, Francesco F. and Franke, Andreas and Hung, Judy and {P{\'e}rez de Isla}, Leopoldo and Kamp, Otto and Kasprzak, Jaroslaw D. and Lancellotti, Patrizio and Marwick, Thomas H. and McCulloch, Marti L. and Monaghan, Mark J. and Nihoyannopoulos, Petros and Pandian, Natesa G. and Pellikka, Patricia A. and Pepi, Mauro and Roberson, David A. and Shernan, Stanton K. and Shirali, Girish S. and Sugeng, Lissa and Ten Cate, Folkert J. and Vannan, Mani A. and Zamorano, Jose Luis and Zoghbi, William A.},
  year = 2012,
  month = jan,
  journal = {Journal of the American Society of Echocardiography},
  volume = {25},
  number = {1},
  pages = {3--46},
  issn = {0894-7317},
  doi = {10.1016/j.echo.2011.11.010},
  urldate = {2026-10-06}
}

@inproceedings{li3DUltrasoundSpine2021,
  title = {{{3D Ultrasound Spine Imaging}} with {{Application}} of {{Neural Radiance Field Method}}},
  booktitle = {2021 {{IEEE International Ultrasonics Symposium}} ({{IUS}})},
  author = {Li, Honggen and Chen, Hongbo and Jing, Wenke and Li, Yuwei and Zheng, Rui},
  year = 2021,
  month = sep,
  pages = {1--4},
  issn = {1948-5727},
  doi = {10.1109/IUS52206.2021.9593917},
  urldate = {2026-10-27}
}

@article{mildenhallNeRFRepresentingScenes2021,
  title = {{{NeRF}}: Representing Scenes as Neural Radiance Fields for View Synthesis},
  shorttitle = {{{NeRF}}},
  author = {Mildenhall, Ben and Srinivasan, Pratul P. and Tancik, Matthew and Barron, Jonathan T. and Ramamoorthi, Ravi and Ng, Ren},
  year = 2021,
  month = dec,
  journal = {Communications of the ACM},
  volume = {65},
  number = {1},
  pages = {99--106},
  issn = {0001-0782},
  doi = {10.1145/3503250},
  urldate = {2026-06-08}
}

@article{nyrnesBloodSpeckleTrackingBased2020,
  title = {Blood {{Speckle-Tracking Based}} on {{High}}--{{Frame Rate Ultrasound Imaging}} in {{Pediatric Cardiology}}},
  author = {Nyrnes, Siri A. and Fadnes, Solveig and Wigen, Morten Smedsrud and Mertens, Luc and Lovstakken, Lasse},
  year = 2020,
  month = apr,
  journal = {Journal of the American Society of Echocardiography},
  volume = {33},
  number = {4},
  pages = {493-503.e5},
  issn = {0894-7317},
  doi = {10.1016/j.echo.2019.11.003},
  urldate = {2025-01-06}
}

@misc{patelDynamicReconstructionUltrasoundDerived2025,
  title = {Dynamic {{Reconstruction}} of {{Ultrasound-Derived Flow Fields With Physics-Informed Neural Fields}}},
  author = {Patel, Viraj and Kreusser, Lisa and Fraser, Katharine},
  year = 2025,
  month = nov,
  number = {arXiv:2511.01804},
  eprint = {2511.01804},
  primaryclass = {cs},
  publisher = {arXiv},
  doi = {10.48550/arXiv.2511.01804},
  urldate = {2025-11-05},
  archiveprefix = {arXiv}
}

@article{poelma3DFlowReconstruction2011,
  title = {{{3D Flow}} Reconstruction Using Ultrasound {{PIV}}},
  author = {Poelma, C. and Mari, J. M. and Foin, N. and Tang, M.-X. and Krams, R. and Caro, C. G. and Weinberg, P. D. and Westerweel, J.},
  year = 2011,
  month = apr,
  journal = {Experiments in Fluids},
  volume = {50},
  number = {4},
  pages = {777--785},
  issn = {1432-1114},
  doi = {10.1007/s00348-009-0781-8},
  urldate = {2026-10-06},
  langid = {english}
}

@misc{pumarolaDNeRFNeuralRadiance2020,
  title = {D-{{NeRF}}: {{Neural Radiance Fields}} for {{Dynamic Scenes}}},
  shorttitle = {D-{{NeRF}}},
  author = {Pumarola, Albert and Corona, Enric and {Pons-Moll}, Gerard and {Moreno-Noguer}, Francesc},
  year = 2020,
  month = nov,
  number = {arXiv:2011.13961},
  eprint = {2011.13961},
  primaryclass = {cs.CV},
  publisher = {arXiv},
  doi = {10.48550/arXiv.2011.13961},
  urldate = {2026-10-06},
  archiveprefix = {arXiv}
}

@article{saricRecommendationsIdentificationMitigation2026,
  title = {Recommendations for the {{Identification}} and {{Mitigation}} of {{Cardiac Ultrasound Artifacts}}: {{A Guideline}} from the {{American Society}} of {{Echocardiography}}},
  shorttitle = {Recommendations for the {{Identification}} and {{Mitigation}} of {{Cardiac Ultrasound Artifacts}}},
  author = {Saric, Muhamed and Sadeghpour, Anita and Alizadeh, Leila and Alizadehasl, Azin and Bertrand, Philippe B. and Billick, Kristen and Chebrolu, Bindu and Faletra, Francesco and Kelsey, Anita and Lang, Roberto and Levine, Robert A. and Pourafkari, Leili and Walling, Steve},
  year = 2026,
  month = may,
  journal = {Journal of the American Society of Echocardiography},
  volume = {39},
  number = {5},
  pages = {435--475},
  issn = {0894-7317},
  doi = {10.1016/j.echo.2026.01.007},
  urldate = {2026-10-08}
}

@inproceedings{sawdayeeOReXObjectReconstruction2023,
  title = {{{OReX}}: {{Object Reconstruction From Planar Cross-Sections Using Neural Fields}}},
  shorttitle = {{{OReX}}},
  booktitle = {Proceedings of the {{IEEE}}/{{CVF Conference}} on {{Computer Vision}} and {{Pattern Recognition}}},
  author = {Sawdayee, Haim and Vaxman, Amir and Bermano, Amit H.},
  year = 2023,
  pages = {20854--20862},
  urldate = {2026-10-27},
  langid = {english}
}

@article{secchiMRImagingAortic2009,
  title = {{{MR}} Imaging of Aortic Coarctation},
  author = {Secchi, F. and Iozzelli, A. and Papini, G. D. E. and Aliprandi, A. and Di Leo, G. and Sardanelli, F.},
  year = 2009,
  month = jun,
  journal = {La radiologia medica},
  volume = {114},
  number = {4},
  pages = {524--537},
  issn = {0033-8362, 1826-6983},
  doi = {10.1007/s11547-009-0386-6},
  urldate = {2025-05-19},
  copyright = {http://www.springer.com/tdm},
  langid = {english}
}

@inproceedings{sitzmannImplicitNeuralRepresentations2020a,
  title = {Implicit {{Neural Representations}} with {{Periodic Activation Functions}}},
  booktitle = {Advances in {{Neural Information Processing Systems}}},
  author = {Sitzmann, Vincent and Martel, Julien and Bergman, Alexander and Lindell, David and Wetzstein, Gordon},
  year = 2020,
  volume = {33},
  pages = {7462--7473},
  publisher = {Curran Associates, Inc.},
  urldate = {2026-11-05}
}

@misc{tancikFourierFeaturesLet2020,
  title = {Fourier {{Features Let Networks Learn High Frequency Functions}} in {{Low Dimensional Domains}}},
  author = {Tancik, Matthew and Srinivasan, Pratul P. and Mildenhall, Ben and {Fridovich-Keil}, Sara and Raghavan, Nithin and Singhal, Utkarsh and Ramamoorthi, Ravi and Barron, Jonathan T. and Ng, Ren},
  year = 2020,
  month = jun,
  number = {arXiv:2006.10739},
  eprint = {2006.10739},
  primaryclass = {cs},
  publisher = {arXiv},
  doi = {10.48550/arXiv.2006.10739},
  urldate = {2025-05-06},
  archiveprefix = {arXiv}
}

@article{voorneveld4DEchoParticleImage2020,
  title = {4-{{D Echo-Particle Image Velocimetry}} in a {{Left Ventricular Phantom}}},
  author = {Voorneveld, Jason and Saaid, Hicham and Schinkel, Christiaan and Radeljic, Nikola and Lippe, Boris and Gijsen, Frank J. H. and {van der Steen}, Antonius F. W. and {de Jong}, Nico and Claessens, Tom and Vos, Hendrik J. and Kenjeres, Sasa and Bosch, Johan G.},
  year = 2020,
  month = mar,
  journal = {Ultrasound in Medicine \& Biology},
  volume = {46},
  number = {3},
  pages = {805--817},
  issn = {0301-5629},
  doi = {10.1016/j.ultrasmedbio.2019.11.020},
  urldate = {2026-10-06}
}

@inproceedings{wangNeuralRenderingStereo2022,
  title = {Neural {{Rendering}} for~{{Stereo 3D Reconstruction}} of~{{Deformable Tissues}} in~{{Robotic Surgery}}},
  booktitle = {Medical {{Image Computing}} and {{Computer Assisted Intervention}} -- {{MICCAI}} 2022},
  author = {Wang, Yuehao and Long, Yonghao and Fan, Siu Hin and Dou, Qi},
  editor = {Wang, Linwei and Dou, Qi and Fletcher, P. Thomas and Speidel, Stefanie and Li, Shuo},
  year = 2022,
  pages = {431--441},
  publisher = {Springer Nature Switzerland},
  address = {Cham},
  doi = {10.1007/978-3-031-16449-1_41},
  isbn = {978-3-031-16449-1},
  langid = {english}
}

@article{wong3DVelocityPressure2025a,
  title = {{{3D}} Velocity and Pressure Field Reconstruction in the Cardiac Left Ventricle via Physics Informed Neural Network from Echocardiography Guided by {{3D}} Color {{Doppler}}},
  author = {Wong, Hong Shen and Chan, Wei Xuan and Mao, Wenbin and Yap, Choon Hwai},
  year = 2025,
  month = may,
  journal = {Computer Methods and Programs in Biomedicine},
  volume = {263},
  pages = {108671},
  issn = {0169-2607},
  doi = {10.1016/j.cmpb.2025.108671},
  urldate = {2026-10-06}
}

@inproceedings{wysockiUltraNeRFNeuralRadiance2024,
  title = {Ultra-{{NeRF}}: {{Neural Radiance Fields}} for {{Ultrasound Imaging}}},
  shorttitle = {Ultra-{{NeRF}}},
  booktitle = {Medical {{Imaging}} with {{Deep Learning}}},
  author = {Wysocki, Magdalena and Azampour, Mohammad Farid and Eilers, Christine and Busam, Benjamin and Salehi, Mehrdad and Navab, Nassir},
  year = 2024,
  month = jan,
  pages = {382--401},
  publisher = {PMLR},
  issn = {2640-3498},
  urldate = {2026-10-26},
  langid = {english}
}

@inproceedings{xieNeuralFieldsVisual2022,
  title = {Neural Fields in Visual Computing and Beyond},
  booktitle = {Computer {{Graphics Forum}}},
  author = {Xie, Yiheng and Takikawa, Towaki and Saito, Shunsuke and Litany, Or and Yan, Shiqin and Khan, Numair and Tombari, Federico and Tompkin, James and Sitzmann, Vincent and Sridhar, Srinath},
  year = 2022,
  volume = {41},
  pages = {641--676},
  publisher = {Wiley Online Library}
}

@article{zhongAssessmentCardiacDysfunction2016,
  title = {Assessment of Cardiac Dysfunction by Dissipative Energy Loss Derived from Vector Flow Mapping},
  author = {Zhong, Yue and Zhu, Wei and Li, Chun-Mei and Rao, Li},
  year = 2016,
  month = jan,
  journal = {Journal of Cardiology},
  volume = {67},
  number = {1},
  pages = {122},
  publisher = {Elsevier},
  issn = {0914-5087},
  doi = {10.1016/j.jjcc.2015.06.004},
  urldate = {2025-05-20},
  langid = {english}
}

@article{caiPhysicsinformedNeuralNetworks2021,
  title = {Physics-Informed Neural Networks ({{PINNs}}) for Fluid Mechanics: A Review},
  author = {Cai, Shengze and Mao, Zhiping and Wang, Zhicheng and Yin, Minglang and Karniadakis, George Em},
  year = 2021,
  journal = {Acta Mechanica Sinica},
  volume = {37},
  pages = {1727--1738}
}

@article{epshteinHighresolution3DFlow2026,
  title = {High-Resolution {{3D}} Flow Reconstruction of Cerebrospinal Fluid Microcirculation Using Physics-Informed Neural Network: Conceptualization and Application to a Large Animal Model},
  author = {Epshtein, Mark and Mekler, Tirosh and Shazeeb, Mohammed Salman and Lindsay, Clifford and Gounis, Matthew J. and Korin, Netanel and Anagnostakou, Vania},
  year = 2026,
  journal = {Computer Methods and Programs in Biomedicine},
  volume = {283},
  pages = {109405},
  doi = {10.1016/j.cmpb.2026.109405}
}

@article{fathiSuperresolutionDenoising4DFlow2020,
  title = {Super-Resolution and Denoising of {{4D-Flow MRI}} Using Physics-Informed Deep Neural Nets},
  author = {Fathi, Mojtaba F. and {Perez-Raya}, Isaac and Baghaie, Ahmadreza and Berg, Philipp and Janiga, Gabor and Arzani, Amirhossein and D'Souza, Roshan M.},
  year = 2020,
  journal = {Computer Methods and Programs in Biomedicine},
  volume = {197},
  pages = {105729},
  doi = {10.1016/j.cmpb.2020.105729}
}

@article{saittaImplicitNeuralRepresentations2024,
  title = {Implicit Neural Representations for Unsupervised Super-Resolution and Denoising of {{4D}} Flow {{MRI}}},
  author = {Saitta, Simone and Carioni, Marcello and Mukherjee, Subhadip and Sch{\"o}nlieb, Carola-Bibiane and Redaelli, Alberto},
  year = 2024,
  journal = {Computer Methods and Programs in Biomedicine},
  volume = {246},
  pages = {108057}
}

@article{vixegeFullvolumeThreecomponentIntraventricular2022,
  title = {Full-Volume Three-Component Intraventricular Vector Flow Mapping by Triplane Color {{Doppler}}},
  author = {Vix{\`e}ge, Florian and Berod, Alain and Courand, Pierre-Yves and Mendez, Simon and Nicoud, Franck and {Blanc-Benon}, Philippe and Vray, Didier and Garcia, Damien},
  year = 2022,
  journal = {Physics in Medicine \& Biology},
  volume = {67},
  number = {9},
  pages = {095004},
  doi = {10.1088/1361-6560/ac62fe}
}
}


\end{document}